\documentclass[runningheads]{llncs}
\usepackage{graphicx}
\usepackage{amsmath,amssymb} 
\pdfoutput=1
\usepackage{color}
\usepackage[width=122mm,left=12mm,paperwidth=146mm,height=193mm,top=12mm,paperheight=217mm]{geometry}
\usepackage{multirow}
\usepackage{wrapfig}

\begin{document}

\pagestyle{headings}
\mainmatter

\title{Video Summarization Using Fully Convolutional Sequence Networks}

\titlerunning{Video Summarization Using Fully Convolutional Sequence Networks}

\authorrunning{Mrigank Rochan, Linwei Ye, Yang Wang}

\author{Mrigank Rochan, Linwei Ye, and Yang Wang\\
\small\texttt{\{mrochan,yel3,ywang\}@cs.umanitoba.ca}}
\institute{University of Manitoba, Canada}

\addtolength{\itemsep}{-0.1in}
\addtolength{\topsep}{-0.07in}
\addtolength{\textfloatsep}{-0.05in}
\addtolength{\intextsep}{-0.05in}
\addtolength{\partopsep}{-0.03in}
\addtolength{\parskip}{-0.02in}

\maketitle

\begin{abstract}
This paper addresses the problem of video summarization. Given an input video, the goal is to select a subset of the frames to create a summary video that optimally captures the important information of the input video. With the large amount of videos available online, video summarization provides a useful tool that assists video search, retrieval, browsing, etc. In this paper, we formulate video summarization as a sequence labeling problem. Unlike existing approaches that use recurrent models, we propose fully convolutional sequence models to solve video summarization. We firstly establish a novel connection between semantic segmentation and video summarization, and then adapt popular semantic segmentation networks for video summarization. Extensive experiments and analysis on two benchmark datasets demonstrate the effectiveness of our models.

\end{abstract}

\section{Introduction}\label{sec:intro}

With the ever-increasing popularity and decreasing cost of video capture devices, the amount of video data has increased drastically in the past few years. Video has become one of the most important form of visual data. Due to the sheer amount of video data, it is unrealistic for humans to watch these videos and identify useful information. According to Cisco Visual Networking Index 2017 \cite{cisco}, it is estimated that it will take around 5 million years for an individual to watch all the videos that are uploaded on the Internet each month in 2021! It is therefore becoming increasingly important to develop computer vision techniques that can enable efficient browsing of the enormous video data. In particular, video summarization has emerged as a promising tool to help cope with the overwhelming amount of video data.

Given an input video, the goal of video summarization is to create a shorter video that captures the important information of the input video. Video summarization can be useful in many real-world applications. For example, in video surveillance, it is tedious and time-consuming for humans to browse through many hours of videos captured by surveillance cameras. If we can provide a short summary video that captures the important information from a long video, it will greatly reduce human efforts required in video surveillance. Video summarization can also provide better user experience in video search, retrieval, and understanding. Since short videos are easier to store and transfer, they can be useful for mobile applications. The summary videos can also help in many downstream video analysis tasks. For example, it is faster to run any other analysis algorithms (e.g. action recognition) on short videos.

In this paper, we consider video summarization as a keyframe selection problem. Given an input video, our goal is to select a subset of the frames to form the summary video. Equivalently, video summarization can also be formulated as a sequence labeling problem, where each frame is assigned a binary label to indicate whether it is selected in the summary video.

Current state-of-the-art methods \cite{zhang16_eccv,mahasseni17_cvpr} consider video summarization as a sequence labeling problem and solve the problem using a variant of recurrent neural networks known as the long short-term memory (LSTM)~\cite{lstm97}. Each time step in the LSTM model corresponds to a frame in the input video. At each time step, the LSTM model outputs a binary value indicating whether this frame is selected in the summary video. The advantage of LSTM is that it can capture long-term structural dependencies among frames. But these LSTM-based models have inherent limitations. The computation in LSTM is usually left-to-right. This means we have to process one frame at a time and each frame must wait until the previous frame is processed. Although bi-directional LSTM (Bi-LSTM)~\cite{schuster97_sp} exists, the computation in either direction of Bi-LSTM still suffers the same problem. Due to this sequential nature, the computation in LSTM cannot be easily parallelized to take full advantage of the GPU hardware. In our work, we propose fully convolutional models that can process all the frames simultaneously, and therefore take the full advantage of GPU parallelization. Our model is partly inspired by some recent work \cite{lea17_cvpr,gehring2017convolutional,BaiTCN2018} in action detection, audio synthesis, and machine translation showing that convolutional models can outperform recurrent models and can take full advantage of GPU parallelization.

\begin{figure*}[t]
	\center
	\includegraphics[width=0.8\textwidth]{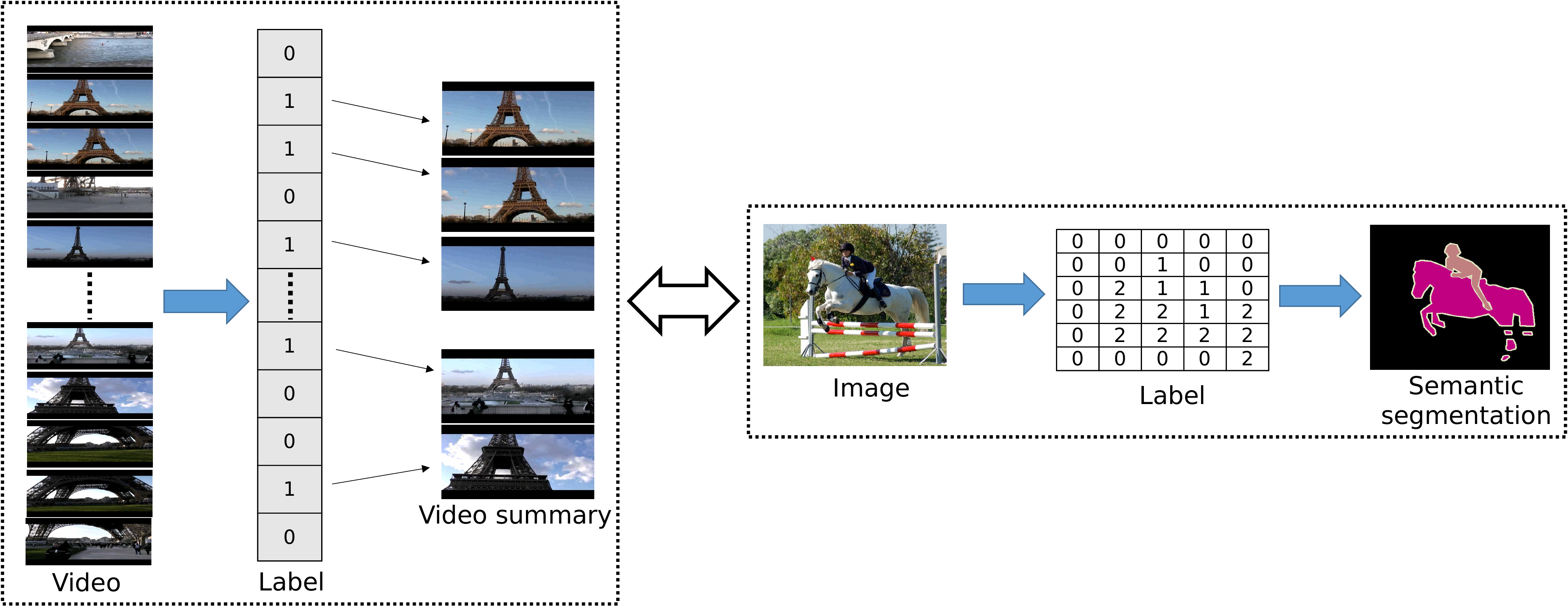}
	\caption{An illustration of the relationship between video summarization and semantic segmentation. (\textit{Left}) In video summarization, our goal is to select frames from an input video to generate the summary video. This is equivalent to assigning a binary label ($0$ or $1$) to each frame in the video to indicate whether the frame is selected for summary. This problem has a close connection with semantic segmentation (\textit{Right}) where the goal is to label each pixel in an image with its class label.} 
	\label{fig:intro}
\end{figure*}

In this paper, we propose to use fully convolutional networks for video summarization. Fully convolutional networks (FCN) \cite{long15_cvpr} have been extensively used in semantic segmentation. Compared with video summarization, semantic segmentation is a more widely studied topic in computer vision. Traditionally, video summarization and semantic segmentation are considered as two completely different problems in computer vision. Our insight is that these two problems in fact share a lot of similarities. In semantic segmentation, the input is a 2D image with 3 color channels (RGB). The output of semantic segmentation is a 2D matrix with the same spatial dimension as the input image, where each cell of the 2D matrix indicates the semantic label of the corresponding pixel in the image. In video summarization, let us assume that each frame is represented as a $K$-dimensional vector. This can be a vector of raw pixel values or a precomputed feature vector. Then the input to video summarization is a 1D image (over temporal dimension) with $K$ channels. The output is a 1D matrix with the same length as the input video, where each element indicates whether the corresponding frame is selected for the summary. In other words, although semantic segmentation and video summarization are two different problems, they only differ in terms of the dimensions of the input (2D vs. 1D) and the number of channels (3 vs. $K$). Figure~\ref{fig:intro} illustrates the relationship between these two tasks. By establishing the connection between these two tasks, we can directly exploit models in semantic segmentation and adapt them for video summarization. In this paper, we develop our video summarization method based on popular semantic segmentation models such as FCN \cite{long15_cvpr}. We call our approach the \emph{Fully Convolutional Sequence Network (FCSN)}.

FCSN is suitable for video summarization due to two important reasons. First, FCSN consist of stack of convolutions whose effective context size grows (though smaller in the beginning) as we go deeper in the network. This allows the network to model the long range complex dependency among input frames that is necessary for video summarization. Second, FCSN is fully convolutional. Compared to LSTM, FCSN allows easier parallelization over input frames.

The contributions of this paper are manifold. (1) To the best of our knowledge, we are the first to propose fully convolutional models for video summarization. (2) We establish a novel connection between two seemingly unrelated problems, namely video summarization and semantic segmentation. We then present a way to adapt popular semantic segmentation networks for video summarization. (3) We propose both supervised and unsupervised fully convolutional models. (4) Through extensive experiments on two benchmark datasets, we show that our model achieves state-of-the-art performance.

\section{Related Work}\label{sec:related}

Given an input video, video summarization aims to produce a shortened version that captures the important information in the video. There are various representations proposed for this problem including video synopsis~\cite{pritch2008nonchronological}, time-lapses~\cite{joshi2015real,kopf2014first,poleg2015egosampling}, montages~\cite{kang2006space,sun2014salient} and storyboards~\cite{gong14_nips,gygli14_eccv,gygli15_cvpr,lee2012discovering,liu2010hierarchical,mahasseni17_cvpr,yang15_iccv,zhang16_cvpr,zhang16_eccv}. Our work is most related to storyboards which select a few representative video frames to summarize key events present in the entire video. Storyboard-based summarization has two types of outputs: keyframes~\cite{gong14_nips,lee2012discovering,liu2010hierarchical} in which certain isolated frames are chosen to form the summary video, and keyshots~\cite{gygli14_eccv,gygli15_cvpr,mahasseni17_cvpr,zhang16_cvpr,zhang16_eccv} in which a set of correlated consecutive frames within a temporal slot are considered for summary generation.

Early work in video summarization mainly relies on hand-crafted heuristics. Most of these approaches are unsupervised. They define various heuristics to represent the importance or representativeness \cite{khosla2013large,kim2014reconstructing_cvpr,lee2012discovering,lu2013story,ngo2003automatic_iccv,Song_2015_CVPR,panda2017collaborative} of the frames and use the importance scores to select representative frames to form the summary video. Recent work has explored supervised learning approaches for video summarization \cite{gong14_nips,gygli14_eccv,gygli15_cvpr,zhang16_cvpr,zhang16_eccv}. These approaches use training data consisting of videos and their ground-truth summaries generated by humans. These supervised learning approaches tend to outperform early work on unsupervised methods, since they can implicitly learn high-level semantic knowledge that is used by humans to generate summaries. 

Recently deep learning methods \cite{zhang16_eccv,mahasseni17_cvpr,sharghi2017query} are gaining popularity for video summarization. The most relevant works to ours are the methods that use recurrent models such as LSTMs~\cite{lstm97}. The intuition of using LSTM is to effectively capture long-range dependencies among video frames which are crucial for meaningful summary generation. Zhang et al.~\cite{zhang16_eccv} consider the video summarization task as a structured prediction problem on sequential data and model the variable-range dependency using two LSTMs. One LSTM is used for video sequences in the forward direction and the other for the backward direction. They further improve the diversity in the subset selection by incorporating a determinantal point process model~\cite{gong14_nips,zhang16_cvpr}. Mahasseni et al.~\cite{mahasseni17_cvpr} propose an unsupervised generative adversarial framework consisting of the summarizer and discriminator. The summarizer is a variational autoencoder LSTM which first selects video frames and then decodes the output for reconstruction. The discriminator is another LSTM network that learns to distinguish between the input video and its reconstruction. They also extend their method to supervised learning by introducing a keyframe regularization. Different from these LSTM-based approaches, we propose fully convolutional sequence models for video summarization. Our work is the first to use fully convolutional models for this problem.

\section{Our Approach}\label{sec:approach}
In this section, we first describe the problem formulation (Sec. \ref{sub:formulation}). We then introduce our fully convolutional sequence model and the learning algorithm (Sec. \ref{sub:fcsn}). Finally, we present an extension of the basic model for unsupervised learning of video summarization (Sec.~\ref{sub:variant}).

\subsection{Problem Formulation}\label{sub:formulation}
Previous work has considered two different forms of output in video summarization: 1) binary labels; 2) frame-level importance scores. Binary label outputs are usually referred to as either keyframes \cite{de2011vsumm,gong14_nips,mundur2006keyframe,zhang16_eccv} or keyshots \cite{gygli14_eccv,gygli15_cvpr,potapov14_eccv,Song_2015_CVPR,zhang16_eccv}. Keyframes consist of a set of non-continuous frames that are selected for the summarization, while keyshots correspond to a set of time-intervals in video where each interval consists of a continuous set of frames. Frame-level importance scores \cite{gygli14_eccv,Song_2015_CVPR} indicate how likely a frame should be selected for the summarization. Existing datasets have ground-truth annotations available in at least one of these two forms. Although frame-level scores provide richer information, it is practically much easier to collect annotations in terms of binary labels. It may even be possible to collect binary label annotations automatically from edited video content online. For example, if we have access to professionally edited summary videos and their corresponding raw videos, we can automatically create annotations in the form of binary labels on frames. In this paper, we focus on learning video summarization from only binary label-based (in particular, keyframe-based) annotations.

Let us consider a video with $T$ frames. We assume each frame has been preprocessed (e.g. by a pretrained CNN) and is represented as a feature vector. We denote the frames in a video as $\{F_1,F_2,F_3,.....,F_T\}$ where $F_i$ is the feature descriptor of the $t$-th ($t\in\{1,2,..,T\}$) frame in the video. Our goal is to assign a binary label (0 or 1) to each of the $T$ frames. The summary video is obtained by combining the frames that are labeled as $1$ (see Fig.~\ref{fig:intro}). We assume access to a training dataset of videos, where each frame has a ground-truth binary label indicating whether this frame should be selected in the summary video.

\subsection{Fully Convolutional Sequence Networks}\label{sub:fcsn} 
Our models are inspired by fully convolutional models used in semantic segmentation. Our models have the following properties. 1) Semantic segmentation models use 2D convolution over 2D spatial locations in an image. In contrast, our models apply 1D convolution across the temporal sequence domain. 2) Unlike LSTM models~\cite{zhang16_eccv} for video summarization that process frames in a sequential order, our models process all frames simultaneously using the convolution operation. 3) Semantic segmentation models usually use an encoder-decoder architecture, where an image is first processed by the encoder to extract features, then the decoder is used to produce the segmentation mask using the encoded features. Similarly, our models can also be interpreted as an encoder-decoder architecture. The encoder is used to process the frames to extract both high-level semantic features and long-term structural relationship information among frames, while the decoder is used to produce a sequence of $0/1$ labels. We call our model the \emph{fully convolutional sequence network (FCSN)}.

Our models mainly consist of temporal modules such as temporal convolution, temporal pooling, and temporal deconvolution. This is analogous to the modules commonly used in semantic segmentation models, such as 2D convolution, 2D pooling, 2D deconvolution. Due to the underlying relationship between video summarization and semantic segmentation, we can easily borrow the network architecture from existing semantic segmentation models when designing FCSN for video summarization. In this section, we describe a FCSN based on a popular semantic segmentation network, namely FCN \cite{long15_cvpr}. We refer to this FCSN as SUM-FCN. It is important to note that FCSN is certainly not limited to this particular network architecture. We can convert almost any existing semantic segmentation models into FCSN for video summarization.
\begin{wrapfigure}{r}{0.5\textwidth}
	\center
	\includegraphics[width=0.5\textwidth]{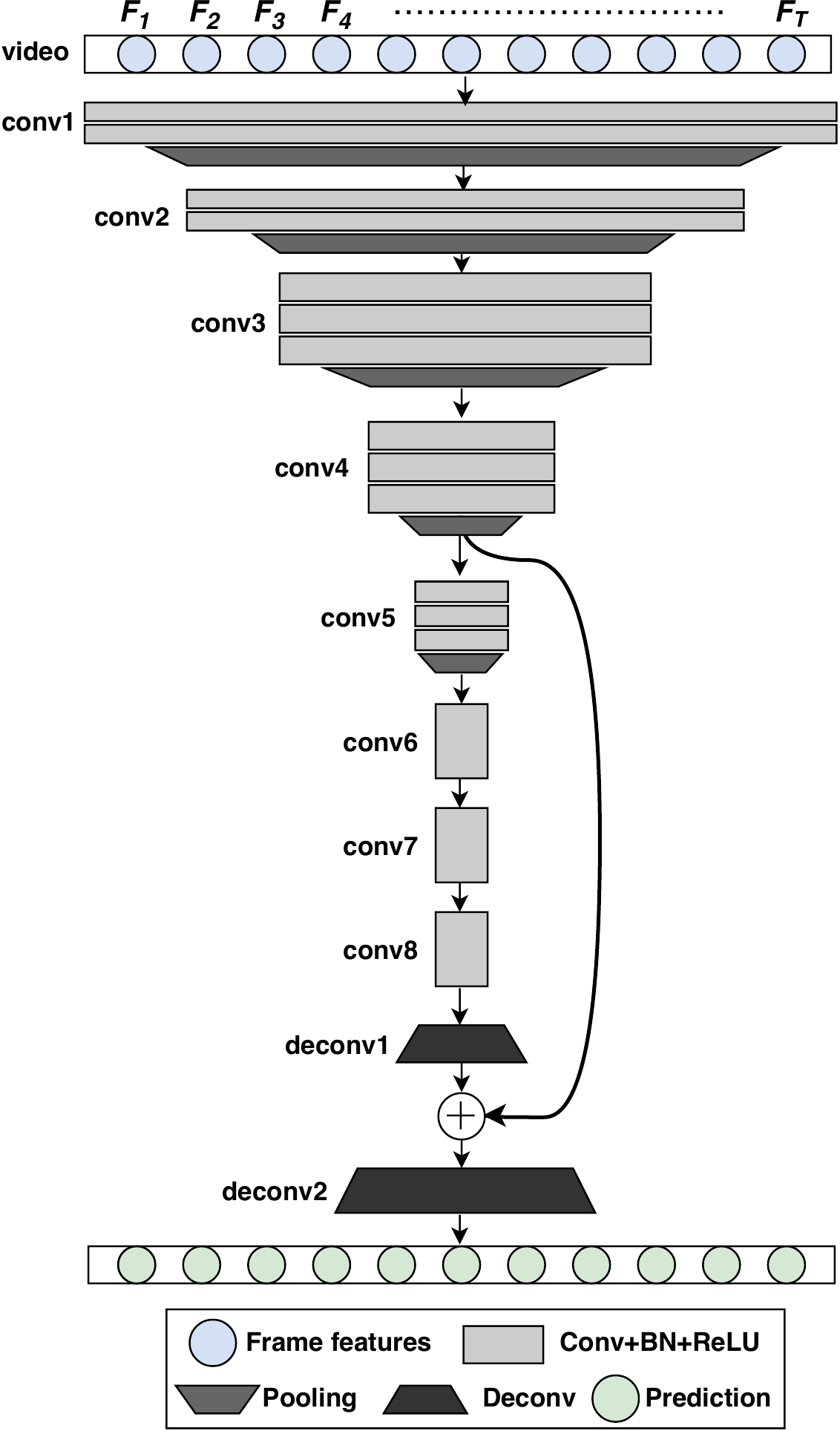}
	\caption{The architecture of SUM-FCN. It is based on the popular semantic segmentation architecture FCN \cite{long15_cvpr}. Unlike FCN, SUM-FCN performs convolution, pooling and deconvolution operation across time.}
	\label{fig:sumfcn}
\end{wrapfigure}

\noindent{\bf SUM-FCN:} FCN \cite{long15_cvpr} is a widely used model for semantic segmentation. In this section, we adapt FCN (in particular, FCN-16) for the task of video summarization. We call the model SUM-FCN. In FCN, the input is an RGB image of shape $m\times n\times 3$ where $m$ and $n$ are height and width of the image respectively. The output/prediction is of shape $m\times n\times C$ where the channel dimension $C$ corresponds to the number of classes. In SUM-FCN, the input is of dimension $1\times T\times D$ where $T$ is the number of frames in a video and $D$ is the dimension of the feature vector of a frame. The output of SUM-FCN is of dimension $1\times T\times C$. Note that the dimension of the output channel is $C=2$ since we need scores corresponding to $2$ classes (keyframe or non-keyframe) for each frame.

Figure \ref{fig:sumfcn} shows the architecture of our SUM-FCN model. We convert all the spatial convolutions in FCN to temporal convolutions. Similarly, spatial maxpooling and deconvolution layers are converted to corresponding temporal counterparts. We organize our network similar to FCN. The first five convolutional layers ($conv1$ to $conv5$) consist of multiple temporal convolution layers where each temporal convolution is followed by a batch normalization and a ReLU activation. We add a temporal maxpooling next to each convolution layer. Each of $conv6$ and $conv7$ consists of a temporal convolution, followed by ReLU and dropout. We also have $conv8$ consisting of a $1 \times 1$ convolution (to produce the desired output channel), batch normalization, and deconvolution operation along the time axis. We then take the output of $pool4$, apply a $1 \times 1$ convolution and batch normalization and then merge (element-wise addition) it with $deconv1$ feature map. This merging corresponds to the skip connection in \cite{long15_cvpr}. Skip connection is widely used in semantic segmentation to combine feature maps at coarse layers with fine layers to produce richer visual features. Our intuition is that this skip connection is also useful in video summarization, since it will help in recovering temporal information required for summarization. Lastly, we apply a temporal deconvolution again and obtain the final prediction of length $T$.

\noindent{\bf Learning:} In keyframe-based supervised setting, the classes (keyframe vs. non-keyframe) are extremely imbalanced since only a small number of frames in an input video are selected in the summary video. This means that there are very few keyframes compared with non-keyframes. A common strategy for dealing with such class imbalance is to use a weighted loss for learning. For the $c$-th class, we define its weight  $w_c=\frac{median\_freq}{freq_c}$, where $freq_c$ is the number of frames with label $c$ divided by the total number of frames in videos where label $c$ is present, and $median\_freq$ is simply the median of the computed frequencies. Note that this class balancing strategy has been used for pixel labeling tasks as well~\cite{eigen15_iccv}.

Suppose we have a training video with $T$ frames. We also have a ground-truth binary label (i.e. number of classes, C = $2$) on each frame of this video. We can define the following loss $\mathcal{L}_{sum}$ for learning:
\begin{equation}
  \label{eq:l_sum}
\mathcal{L}_{sum} = -\frac{1}{T}\sum_{t=1}^{T} w_{c_t}\log\Big(\frac{\exp(\phi_{t,c_{t}})}{\sum_{c=1}^{C}\exp(\phi_{t,c})}\Big)
\end{equation}
where $c_{t}$ is the ground-truth label of the $t$-th frame. $\phi_{t,c}$ and $w_c$ indicate the score of predicting the $t$-th frame as the $c$-th class and the weight of class $c$, respectively.

\subsection{Unsupervised SUM-FCN}\label{sub:variant}
In this section, we present an extension of the SUM-FCN model. We develop an unsupervised variant (called SUM-FCN$_{unsup}$) of SUM-FCN to learn video summarization from a collection of raw videos without their ground-truth summary videos.

Intuitively, the frames in the summary video should be visually diverse~\cite{zhang16_eccv,mahasseni17_cvpr}. We use this property of video summarization to design SUM-FCN$_{unsup}$. We develop SUM-FCN$_{unsup}$ by explicitly encouraging the model to generate summary videos where the selected frames are visually diverse. In order to enforce this diversity, we make the following changes to the decoder of SUM-FCN. We first select $Y$ frames (i.e. keyframes) based on the prediction scores from the decoder. Next, we apply a $1\times1$ convolution to the decoded feature vectors of these keyframes to reconstruct their original feature representations. We then merge the input frame-level feature vectors of these selected $Y$ keyframes using a skip connection. Finally, we use a $1\times1$ convolution to obtain the final reconstructed features of the $Y$ keyframes such that each keyframe feature vector is of the same dimension as its corresponding input frame-level feature vector.

We use a repelling regularizer \cite{zhao2016energy} $\mathcal{L}_{div}$ to enforce diversity among selected keyframes. We define $\mathcal{L}_{div}$ as the mean of the pairwise similarity between the selected $Y$ keyframes:
\begin{equation}
  \label{eq:l_div}
\mathcal{L}_{div} = \frac{1}{|Y|(|Y|-1)}\sum_{t\in Y}\sum_{\substack{t'\in Y, t'\neq t}}d(f_t,f_{t'}),\;\textrm{where}\;(f_t,f_{t'})=\frac{f_t^T {f_{t'}}}{\Vert f_t \Vert _2 \Vert f_{t'} \Vert _2}
\end{equation}
where $f_t$ is the reconstructed feature vector of the frame $t$. Ideally, a diverse subset of frames will lead to a lower value of $\mathcal{L}_{div}$.

We also introduce a reconstruction loss $\mathcal{L}_{recon}$ that computes the mean squared error between the reconstructed features and the input feature vectors of the keyframes. The final learning objective of SUM-FCN$_{unsup}$ becomes $\mathcal{L}_{div}+\mathcal{L}_{recon}$. Since this objective does not require ground-truth summary videos, SUM-FCN$_{unsup}$ is an unsupervised approach.

It is worth noting that SUM-FCN will implicitly achieve diversity to some extent because it is supervised. SUM-FCN learns to mimic the ground-truth human annotations. Presumably, the ground-truth summary videos (annotated by humans) have diversity among the selected frames, since humans are unlikely to annotate two very similar frames as keyframes.
\section{Experiments}\label{sec:experiments}
In this section, we first introduce the datasets in Sec.~\ref{sub:dataset}. We then discuss the implementation details and setup in Sec.~\ref{sub:imp_details}. Lastly, we present the main results in Sec.~\ref{sub:results_fcn} and additional ablation analysis in Sec.~\ref{sec:analysis}.
\subsection{Datasets}\label{sub:dataset}

We evaluate our method on two benchmark datasets: SumMe \cite{gygli14_eccv} and TVSum \cite{Song_2015_CVPR}. The SumMe dataset is a collection of 25 videos that cover a variety of events (e.g. sports, holidays, etc.). The videos in SumMe are 1.5 to 6.5 minutes in length. The TVSum dataset contains 50 YouTube videos of 10 different categories (e.g. making sandwich, dog show, changing vehicle tire, etc.) from the TRECVid Multimedia Event Detection (MED) task \cite{smeaton2006evaluation_mir}. The videos in this dataset are typically 1 to 5 minutes in length. 

Since training a deep neural network with small annotated datasets is difficult, previous work~\cite{zhang16_eccv} has proposed to use additional videos to augment the datasets. Following \cite{zhang16_eccv}, we use 39 videos from the YouTube dataset \cite{de2011vsumm} and 50 videos from the Open Video Project (OVP) dataset \cite{de2011vsumm,ovp} to augment the training data. In the YouTube dataset, there are videos consisting of news, sports and cartoon. In the OVP dataset, there are videos of different genres such as documentary. These datasets are diverse in nature and come with different types of annotations. We discuss in Sec.~\ref{sub:imp_details} on how we handle different formats of ground-truth annotations.

\subsection{Implementation Details and Setup}\label{sub:imp_details}

\noindent {\bf Features:} Following \cite{zhang16_eccv}, we uniformly downsample the videos to $2$ fps. Next, we take the output of the $pool5$ layer in the pretrained GoogleNet \cite{szegedy15_cvpr} as the feature descriptor for each video frame. The dimension of this feature descriptor is 1024. Note that our model can be used with any feature representation. We can even use our model with video-based features (e.g. C3D~\cite{tran15_iccv}). We use GoogleNet features mainly because they are used in previous work~\cite{zhang16_eccv,mahasseni17_cvpr} and will allow fair comparison in the experiments.

\noindent {\bf Ground-truth:} Since different datasets provide the ground-truth annotations in various format, we follow \cite{gong14_nips,zhang16_eccv} to generate the single set of ground-truth keyframes (small subset of isolated frames) for each video in the datasets. These keyframe-based summaries are used for training.

To perform fair comparison with state-of-the-art methods (see Evaluation Metrics below), we need summaries in the form of keyshots (interval-based subset of frames \cite{gygli14_eccv,gygli15_cvpr,zhang16_eccv}) in both the final generated predictions and the ground-truth annotations for test videos. For the SumMe dataset, ground-truth annotations are available in the form of keyshots, so we use these ground-truth summaries directly for evaluation. However, keyshot annotations are missing from the TVSum dataset. TVSum provides frame-level importance scores annotated by multiple users. To convert importance scores to keyshot-based summaries, we follow the procedure in \cite{zhang16_eccv} which includes the following steps: 1) temporally segment a video using KTS \cite{potapov14_eccv} to generate disjoint intervals; 2) compute average interval score and assign it to each frame in the interval; 3) rank the frames in the video based on their scores; 4) apply the knapsack algorithm \cite{Song_2015_CVPR} to select frames so that the total length is under certain threshold, which results in the keyshot-based ground-truth summaries of that video. We use this keyshot-based annotation to get the keyframes for training by selecting the frames with the highest importance scores~\cite{zhang16_eccv}. Note that both the keyframe-based and keyshot-based summaries are represented as $0/1$ vector of length equal to the number of frames in the video. Here, a label $0/1$ represents whether a frame is selected in the summary video. Table \ref{table:datasets_gt} illustrates the ground-truth (training and testing) annotations and their conversion for different datasets. 
\begin{table}[ht]
\caption{Ground-truth (GT) annotations used during training and testing for different datasets. \textsuperscript{\ddag}We convert frame-level importance scores from multiple users to single keyframes as in \cite{Song_2015_CVPR,zhang16_eccv}. \textsuperscript{\dag}We follow \cite{zhang16_eccv} to convert multiple frame-level scores to keyshots. \textsuperscript{\S}Following \cite{gong14_nips,zhang16_eccv}, we generate one set of keyframes for each video. Note that the YouTube and OVP datasets are only used to supplement the training data (as in \cite{zhang16_eccv,mahasseni17_cvpr}), so we do not test our methods on them}
\begin{center}
\begin{tabular}{c|c|c|c}
	\hline
	Dataset & \# annotations & Training GT & Testing GT\\
    \hline
    SumMe & 15-18 & frame-level scores\textsuperscript{\ddag}& keyshots \\
     TVSum & 20 & frame-level scores\textsuperscript{\ddag} & frame-level scores \dag\\
     YouTube & 5 & keyframes\textsuperscript{\S} & - \\
     OVP & 5 & keyframes\textsuperscript{\S} & - \\
    \hline
\end{tabular}
\end{center}
\label{table:datasets_gt}
\end{table}

\noindent {\bf Training and Optimization:} 
We use keyframe-based ground-truth annotations during training. We first concatenate the visual features of each frame. For a video with $T$ frames, we will have an input of dimension $1 \times T \times 1024$ to the neural network. We also uniformly sample frames from each video such that we end up with $T=320$. This sampling is similar to the fixed size cropping in semantic segmentation, where training images are usually resized to have the same spatial size. Note that our proposed model, SUM-FCN, can also effectively handle longer and variable length videos (see Sec.~\ref{sec:analysis}).

During training, we set the learning rate to $10^{-3}$, momentum to $0.9$, and batch size to $5$. Other than using the pretrained GoogleNet to extract frame features, the rest of the network is trained end-to-end using stochastic gradient descent (SGD) optimizer.

\noindent {\bf Testing:}
At test time, a uniformly sampled test video with $T=320$ frames is forwarded to the trained model to obtain an output of length $320$. Then this output is scaled to the original length of the video using nearest-neighbor. For simplicity, we use this strategy to handle test videos. But since our model is fully convolutional, it is not limited to this particular choice of video length. In Sec.~\ref{sec:analysis}, we experiment with sampling the
videos to a longer length. We also experiment with directly operating on original non-sampled (variable length) videos in Sec.~\ref{sec:analysis}.

We follow \cite{zhang16_eccv,mahasseni17_cvpr} to convert predicted keyframes to keyshots so that we can perform fair comparison with other methods. We first apply KTS \cite{potapov14_eccv} to temporally segment a test video into disjoint intervals. Next, if an interval contains a keyframe, we mark all the frames in that interval as $1$ and we mark $0$ to all the frames in intervals that have no keyframes. This results in keyshot-based summary for the video. To minimize the number of generated keyshots, we rank the intervals based on the number of keyframes in intervals divided by their lengths, and finally apply knapsack algorithm \cite{Song_2015_CVPR} to ensure that the produced keyshot-based summary is of maximum $15\%$ in length of the original test video.   

\noindent {\bf Evaluation Metrics:} Following \cite{zhang16_eccv,mahasseni17_cvpr}, we use a keyshot-based evaluation metric. For a given video $V$, suppose $S_O$ is the generated summary and $S_G$ is the ground-truth summary. We calculate the precision ($P$) and recall ($R$) using their temporal overlap:\\
\begin{equation}
P=\frac{|S_O \cap S_G|}{|S_O|}, R=\frac{|S_O \cap S_G|}{|S_G|}
\end{equation}

Finally, we use the F-score $F=(2P \times  R)/(P + R) \times 100\%$ as the evaluation metric. We follow the standard approach described in \cite{Song_2015_CVPR,gygli15_cvpr,zhang16_eccv} to calculate the metric for videos that have multiple ground-truth summaries. 

\noindent\textbf{Experiment Settings}: Similar to previous work~\cite{zhang16_cvpr,zhang16_eccv}, we evaluate and compare our method under the following three different settings.

1. \textit{Standard Supervised Setting}: This is the conventional supervised learning setting where training, validation and test data are drawn (such that they do not overlap) from the same dataset. We randomly select $20\%$ for testing and leave the rest $80\%$ for training and validation. Since the data is randomly splitted, we repeat the experiment over multiple random splits and report the average F-score performance.

2. \textit{Augmented Setting}: For a given dataset, we randomly select $20\%$ data for testing and leave the rest $80\%$ for training and validation. In addition, we use the other three datasets to augment the training data. For example, suppose we are evaluating on the SumMe dataset, we will then have $80\%$ of SumMe videos combined with all the videos in the TVSum, OVP, and YouTube dataset for training. Likewise, if we are evaluating on TVSum, we will have $80\%$ of TVSum videos combined with all the videos in SumMe, OVP, and YouTube for training. Similar to the standard supervised setting, we run the experiment over multiple random splits and use the average F-score for comparison.

The idea of increasing the size of training data by augmenting with other datasets is well-known in computer vision. This is usually referred as data augmentation. Recent methods \cite{zhang16_eccv,mahasseni17_cvpr} show that data augmentation improves the performance. Our experimental results show similar conclusion.

3. \textit{Transfer Setting}: This is a challenging supervised setting introduced by Zhang et al. \cite{zhang16_cvpr,zhang16_eccv}. In this setting, the model is not trained using the videos from the given dataset. Instead, the model is trained on other available datasets and tested on the given dataset. For instance, if we are evaluating on the SumMe dataset, we will train the model using videos in the TVSum, OVP, and YouTube datasets. We then use the videos in the SumMe dataset only for evaluation. Similarly, when evaluating on TVSum, we will train on videos from SumMe, OVP, YouTube, and then test on the videos in TVSum. This setting is particularly relevant for practical applications. If we can achieve good performance under this setting, it means that we can perform video summarization in the wild. In other words, we will be able to generate good summaries for videos from domains in which we do not have any related annotated videos during training.

\subsection{Main Results and Comparisons}\label{sub:results_fcn}
We compare the performance of our approach (SUM-FCN) with prior methods on the SumMe dataset in Table \ref{table:summe}. Our method outperforms other state-of-the-art approaches by a large margin.

\begin{table}[h]
	\caption{Comparison of summarization performance (F-score) between SUM-FCN and other approaches on the SumMe dataset under different settings}
	\begin{center}
		\begin{tabular}{c|c|c|c|c}
			\hline
			Dataset & Method & Standard Supervised & Augmented & Transfer \\
			\hline
			\multirow{7}{*}{SumMe}& Gygli et al. \cite{gygli14_eccv} & 39.4 & -- & --\\
			& Gygli et al. \cite{gygli15_cvpr} & 39.7 & -- & --\\
			& Zhang et al. \cite{zhang16_cvpr} & 40.9 & 41.3 & 38.5\\
			& Zhang et al. \cite{zhang16_eccv} (vsLSTM) & 37.6 & 41.6 & 40.7\\
			& Zhang et al. \cite{zhang16_eccv} (dppLSTM) & 38.6 & 42.9 & 41.8\\
			& Mahasseni et al. \cite{mahasseni17_cvpr} (supervised) & 41.7 & 43.6 & --\\
			& Li et al. \cite{li2017_tip} & 43.1 & -- & -- \\ \cline{2-5}
			& SUM-FCN (ours) & {\bf 47.5} & {\bf 51.1} & {\bf 44.1}\\
			\hline
		\end{tabular}
	\end{center}
	\label{table:summe}
\end{table}

Table \ref{table:tvsum} compares the performance of our method with previous approaches on the TVSum dataset. Again, our method achieves state-of-the-art performance. In the \textit{standard supervsised} setting, we outperform other approaches. In the \textit{augmented} and \textit{transfer} settings, our performance is comparable to other state-of-the-art. Note that Zhang et al. \cite{zhang16_eccv} (vsLSTM) use frame-level importance scores and Zhang et al. \cite{zhang16_eccv} (dppLSTM) use both keyframe-based annotation and frame-level importance scores. But we only use keyframe-based annotation in our method. Previous method \cite{zhang16_eccv} has also shown that frame-level importance scores provide richer information than binary labels. Therefore, the performance of our method on TVSum is very competitive, since it does not use frame-level importance scores during training.

\begin{table}[h]
	\caption{Performance (F-score) of SUM-FCN and other approaches on the TVSum dataset. \textsuperscript{\dag}Zhang et al. \cite{zhang16_eccv} (vsLSTM) use frame-level importance scores. \textsuperscript{\ddag}Zhang et al. \cite{zhang16_eccv} (dppLSTM) use both frame-level importance scores and keyframes in their method. Different from these two methods, our method only uses keyframe-based annotations}
	\begin{center}
		\begin{tabular}{c|c|c|c|c}
			\hline
			Dataset & Method & Standard Supervised & Augmented & Transfer \\
			\hline
			\multirow{5}{*}{TVSum}& Zhang et al. \cite{zhang16_eccv} (vsLSTM) & 54.2 & 57.9 & 56.9\textsuperscript{\dag}\\
			& Zhang et al. \cite{zhang16_eccv} (dppLSTM) & 54.7 & 59.6 & {\bf 58.7}\textsuperscript{\ddag}\\    						
			& Mahasseni et al. \cite{mahasseni17_cvpr} (supervised) & 56.3 & {\bf 61.2} & -- \\
			& Li et al. \cite{li2017_tip} & 52.7 & -- & --\\\cline{2-5}
			& SUM-FCN (ours) & {\bf 56.8} & 59.2 & 58.2 \\
			\hline
		\end{tabular}
	\end{center}
	\label{table:tvsum}
\end{table}

\subsection{Analysis}\label{sec:analysis}
In this section, we present additional ablation analysis on various aspects of our model.

\noindent{\bf Unsupervised SUM-FCN$_{unsup}$:} Table \ref{table:fcn_unsupervised} compares the performance of SUM-FCN$_{unsup}$ with the other unsupervised methods in the literature. SUM-FCN$_{unsup}$ achieves the state-of-the-art performance on both the datasets. These results suggest that our fully convolutional sequence model can effectively learn how to summarize videos in an unsupervised way. This is very appealing since collecting labeled training data for video summarization is difficult.
\begin{table}[h]
	\caption{Performance (F-score) comparison of SUM-FCN$_{unsup}$ with state-of-the-art unsupervised methods}
	\begin{center}
		\begin{tabular}{c|c|c|c|c|c|c|c}
			\hline
			Dataset & \cite{de2011vsumm} & \cite{li2010multi} & \cite{khosla2013large}  & \cite{Song_2015_CVPR} & \cite{zhao2014quasi} & \cite{mahasseni17_cvpr} & SUM-FCN$_{unsup}$\\
			\hline
			SumMe& 33.7 & 26.6 & -- & 26.6 & -- & 39.1& \textbf{41.5}\\\hline
			TVSum& -- &--& 36.0 & 50.0 & 46.0 & 51.7& \textbf{52.7}\\
			\hline
		\end{tabular}
	\end{center}
	\label{table:fcn_unsupervised}
\end{table}

\noindent{\bf SUM-DeepLab:} To demonstrate the generality of FCSN, we also adapt DeepLab \cite{chen2017deeplab_pami} (in particular, DeepLabv2 (VGG16) model), another popular semantic segmentation model, for video summarization. We call this network SUM-DeepLab. The DeepLab model has two important features: 1) dilated convolution; 2) spatial pyramid pooling. In SUM-DeepLab, we similarly perform temporal dilated convolution and temporal pyramid pooling.

Table \ref{table:deeplab} compares SUM-DeepLab with SUM-FCN on the SumMe and TVSum datasets under different settings. SUM-DeepLab achieves better performance on SumMe in all settings. On TVSum, the performance of SUM-DeepLab is better than SUM-FCN in the \textit{standard supervised} setting and is comparable in the other two settings. 

We noticed that SUM-DeepLab performs slightly worse than SUM-FCN in some settings (e.g. \textit{transfer} setting of TVSum). One possible explanation is that the bilinear upsampling layer in DeepLab may not be the best choice. Unlike semantic segmentation, a smooth labeling (due to bilinear upsampling) is not necessarily desirable in video summarization. In other words, the bilinear upsampling may result in a sub-optimal subset of keyframes. In order to verify this, we replace the bilinear upsampling layers of SUM-DeepLab with learnable deconvolution layers (also used in SUM-FCN) and examine the performance of this modified SUM-DeepLab in the \textit{transfer} setting. The performance of SUM-DeepLab improves as a result of this simple modification. In fact, SUM-DeepLab now achieves the state-of-the-art performance on the \textit{transfer} setting on TVSum as well (see the last column in Table \ref{table:deeplab}).
\begin{table}[h]
	\caption{Performance (F-score) of SUM-DeepLab in different settings. We include the performance of SUM-FCN (taken from Table \ref{table:summe} and Table \ref{table:tvsum}) in brackets. We also replace the bilinear upsampling with learnable deconvolutional layer and report the result in the transfer setting (last column)}
	\begin{center}
		\begin{tabular}{c|c|c|c|c}
			\hline
			Dataset & Standard Supervised & Augmented & Transfer  & Transfer (deconv)\\
			\hline
			SumMe& \textbf{48.8} (47.5) & 50.2 (51.1) & \textbf{45.0} (44.1) & \textbf{45.1}\\\hline
			TVSum& \textbf{58.4} (56.8) & 59.1 (59.2) & 57.4 (58.2) & \textbf{58.8}\\
			\hline
		\end{tabular}
	\end{center}
	\label{table:deeplab}
\end{table}

\noindent{\bf Length of Video:} We also perform experiments to analyze the performance of our models on longer-length videos. Again, we select the challenging \textit{transfer} setting to evaluate the models when the videos are uniformly sampled to $T$=$640$ frames. Table \ref{table:length} (first two columns) shows the results of our models for this case. Compared with $T=320$ (shown in brackets in Table~\ref{table:length}), the performance with $T=640$ is similar. This shows that the video length is not an issue for our proposed fully convolutional models.
\begin{table}[h]
	\caption{Performance (F-score) of our models on longer-length videos (i.e. $T$=$640$) and original (i.e. variable length) videos in the \textit{transfer} data setting. In brackets, we show the performance of our model for $T$=$320$ (obtained from Tables \ref{table:summe}, \ref{table:tvsum}, and \ref{table:deeplab})}
	\begin{center}
          \begin{tabular}{c|c|c||c}
            \hline
            \multirow{2}{*}{Dataset} & SUM-FCN & SUM-DeepLab & SUM-FCN\\
            &T=640 (T=320) & T=640 (T=320) & variable length\\
            \hline
            SumMe & 45.6 (44.1) & 44.5 (45.0) & 46.0\\
            TVSum & 57.4 (58.2) & 57.2 (57.4) & 56.7\\
            \hline
          \end{tabular}
	\end{center}
	\label{table:length}
\end{table}

As mentioned earlier, the main idea behind uniformly sampling videos is to mimic the prevalent cropping strategy in semantic segmentation. Nevertheless, since our model is fully convolutional, it can also directly handle variable length videos. The last column of Table~\ref{table:length} shows the results of applying SUM-FCN (in the \textit{transfer} setting) without sampling videos. The performance is comparable (even higher on SumMe) to the results of sampling videos to a fixed length. 

\begin{figure}
	\centering
	\includegraphics[width=0.95\textwidth]{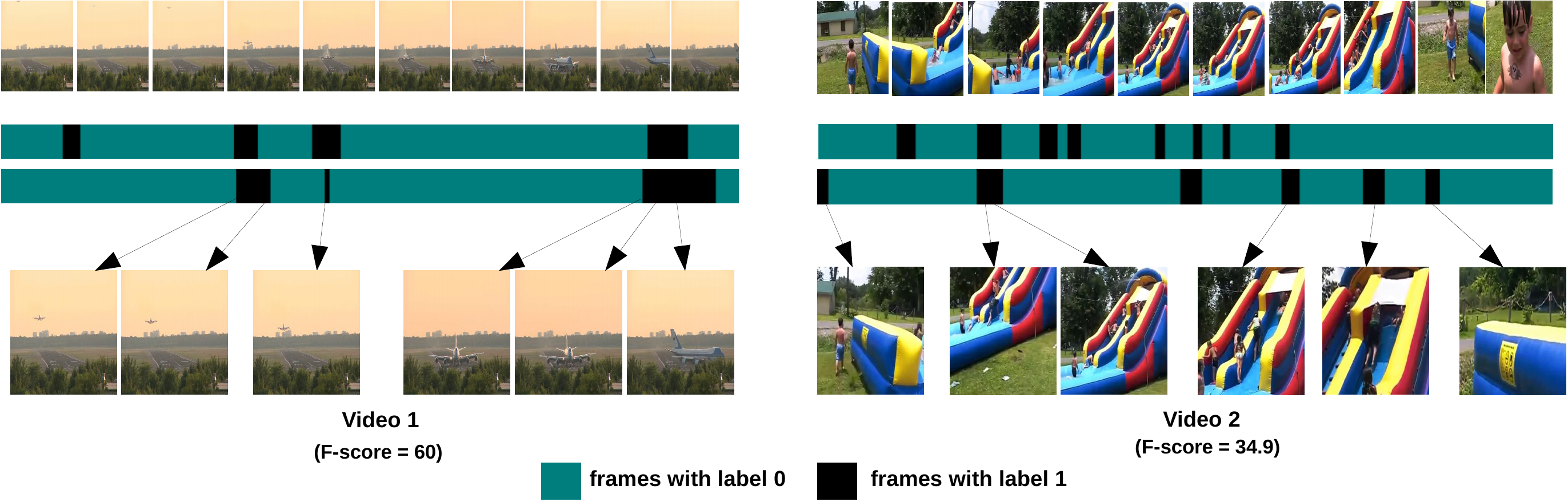}
	\caption{Example summaries for two videos in the SumMe \cite{gygli14_eccv} dataset. The black bars on the green background show the frames selected to form the summary video. For each video, we show the ground-truth (\textit{top bar}) and the predicted labels (\textit{bottom bar}).}
	\label{fig:qualitative}
\end{figure}

\noindent{\bf Qualitative Results:} In Fig. \ref{fig:qualitative}, we show example video summaries (good and poor) produced by SUM-FCN on two videos in the SumMe \cite{gygli14_eccv} dataset.
\section{Conclusion}\label{sec:conclude}

We have introduced fully convolutional sequence networks (FCSN) for video summarization. Our proposed models are inspired by fully convolutional networks in semantic segmentation. In computer vision, video summarization and semantic segmentation are often studied as two separate problems. We have shown that these two seemingly unrelated problems have an underlying connection. We have adapted popular semantic segmentation networks for video summarization. Our models achieve very competitive performance in comparison with other supervised and unsupervised state-of-the-art approaches that mainly use LSTMs. We believe that fully convolutional models provide a promising alternative to LSTM-based approaches for video summarization. Finally, our proposed method is not limited to FCSN variants that we introduced. Using similar strategies, we can convert almost any semantic segmentation networks for video summarization. As future work, we plan to explore more recent semantic segmentation models and develop their counterpart models in video summarization.\\

\noindent \textbf{Acknowledgments:} This work was supported by NSERC, a University of Manitoba Graduate Fellowship, and the University of Manitoba GETS program. We thank NVIDIA for donating some of the GPUs used in this work.


\bibliographystyle{splncs}
\bibliography{wang}
\end{document}